\documentclass[10pt,twocolumn,letterpaper]{article}

\usepackage[pagenumbers]{iccv} % To force page numbers, e.g. for an arXiv version

\definecolor{iccvblue}{rgb}{0.21,0.49,0.74}
\usepackage[pagebackref,breaklinks,colorlinks,allcolors=iccvblue]{hyperref}
\usepackage{graphicx,rotating,multirow,makecell,amssymb,siunitx,microtype,pifont}
\newcommand{\cmark}{\ding{51}}
\newcommand{\xmark}{\ding{55}}
\title{Boosting Multi-View Stereo with Depth Foundation Model \\ in the Absence of Real-World Labels}

\author{{Jie Zhu{$^{1}$}} \quad Bo Peng{$^{1}$} \quad Zhe Zhang{$^{2}$} \quad Bingzheng Liu{$^{1}$} \quad Jianjun Lei{$^{1}$}\thanks{Corresponding author.}\\
	\normalsize
	$^{1}$\	Tianjin University ~~ $^{2}$\, Tianjin University of Commerce\\
	\normalsize
	{\tt\small  \{jzhu98, bpeng, bzliu, jjlei\}@tju.edu.cn \quad zhangz@tjcu.edu.cn}
}

\begin{document}
\maketitle
\begin{abstract}
Learning-based Multi-View Stereo (MVS) methods have made remarkable progress in recent years. However, how to effectively train the network without using real-world labels remains a challenging problem. In this paper, driven by the recent advancements of vision foundation models, a novel method termed DFM-MVS, is proposed to leverage the depth foundation model to generate the effective depth prior, so as to boost MVS in the absence of real-world labels. Specifically, a depth prior-based pseudo-supervised training mechanism is developed to simulate realistic stereo correspondences using the generated depth prior, thereby constructing effective supervision for the MVS network. Besides, a depth prior-guided error correction strategy is presented to leverage the depth prior as guidance to mitigate the error propagation problem inherent in the widely-used coarse-to-fine network structure. Experimental results on DTU and Tanks \& Temples datasets demonstrate that the proposed DFM-MVS significantly outperforms existing MVS methods without using real-world labels. 
\end{abstract}    
\section{Introduction}
\label{sec1}

As a long-standing and fundamental task in 3D computer vision, multi-view stereo (MVS) has attracted significant interest in recent years. Owing to the ability to recover 3D models of the observed scene with calibrated multi-view images, MVS plays a vital role in various applications, such as robot navigation, autonomous driving, and augmented reality~\cite{tutorial}. Traditionally, MVS has been tackled by computing dense stereo correspondences with hand-crafted similarity metrics and regularization strategies~\cite{colmap, acmh, acmmp}. With the development of deep learning, learning-based MVS methods~\cite{mvsnet, casmvsnet, mvsformer++} have made great progress and demonstrated superior performance over the traditional counterparts. Nevertheless, most learning-based methods rely on real-world depth labels collected from 3D scanners, such as structured-light cameras and LiDAR sensors, to provide supervision for networks. In practice, the label collecting process is time-consuming and labor-intensive, which poses a significant challenge for their applicability and scalability.

\begin{figure}
	\centering
	\includegraphics[width=1\linewidth]{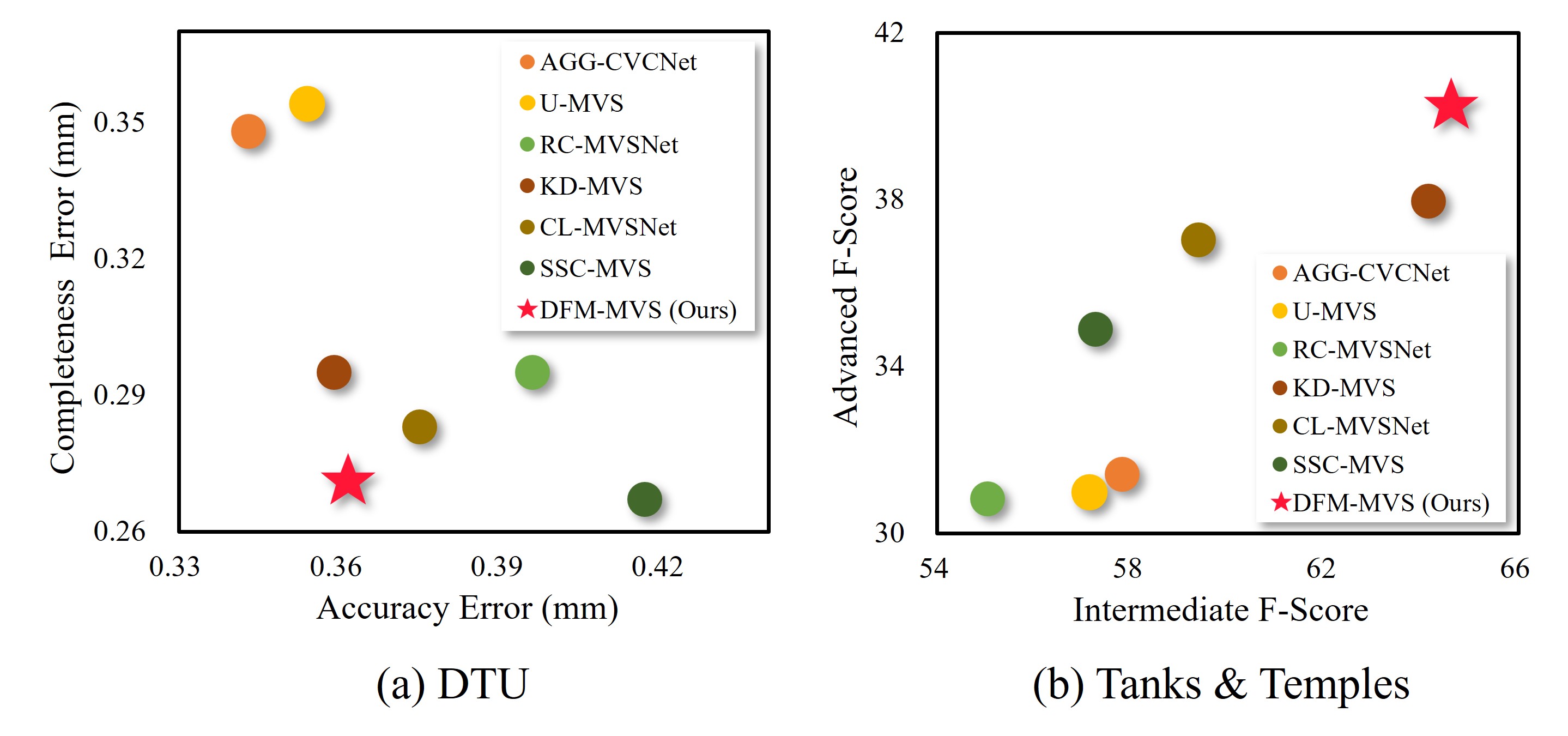}
	\vspace{-0.6cm}
	\caption{Performance comparison of state-of-the-art MVS methods without requiring real-world labels on (a) DTU (\textbf{lower is better}) and (b) Tanks \& Temples (\textbf{higher is better}).}
	\label{pre}
	\vspace{-0.7cm}
\end{figure}

To tackle this challenge, more and more researchers are exploring alternative approaches to train networks without using real-world depth labels~\cite{unsupmvs, jcsda, ssc-mvs}. For example, some methods transform MVS into an image reconstruction problem, and impose the photometric consistency constraint between the reconstructed image and original image to train networks, thus alleviating dependence on depth labels ~\cite{rcmvsnet, kdmvs, clmvs}. Besides, Zhu \etal~\cite{ssc-mvs} employed a superpixel segmentation algorithm to partition the image into distinct regions, and then assigned random depth values to each region, thereby generating the pseudo depth for supervision. While this method facilitates more effective network training compared to the photometric consistency constraint, the manually-designed pseudo depth suffers from suboptimal quality, which restricts the accuracy and applicability of supervision. As a result, further exploration is needed to develop more robust supervision mechanisms in the absence of real-world labels. Moreover, current state-of-the-art MVS methods~\cite{umvs, selfsup-cvp} typically adopt the coarse-to-fine network structure, where the low-resolution depth is first predicted at the coarse scale, and then progressively refined in fine scales to improve resolution and accuracy ~\cite{casmvsnet, cvpmvsnet}. However, the coarse scale often suffers from the inevitable errors due to the inherent limitations, such as loss of texture details from the low resolution. These errors may easily propagate and accumulate in finer scales, thereby affecting the final depth estimation accuracy. Therefore, exploring an effective strategy to alleviate the error propagation problem is crucial to improve the overall reconstruction performance. 

Recently, with the revolution driven by foundation models in computer vision~\cite{clip, sam, stablediff}, there have been several depth foundation models emerging, such as MiDaS~\cite{midas}, Depth Anything~\cite{dav1}, and Depth Anything V2~\cite{dav2}. Although these depth foundation models struggle to generate point clouds for the whole 3D scene due to scale inconsistency across multiple views, they demonstrate strong capability of monocular relative depth estimation in diverse scenarios. Notably, Depth Anything V2 stands out by predicting depth with remarkable real-world characteristics, such as smooth surfaces and sharp boundaries, without relying on any real-world labeled data during training. Motivated by this, we propose to exploit the power of Depth Anything V2 to address the above-mentioned problems, including the unsatisfactory pseudo-supervision and error propagation issue, thus boosting MVS in the absence of real-world labels. As shown in \cref{pre}, the proposed method achieves significant performance improvements over the state-of-the-art methods without requiring real-world labels. The main contributions of this paper are as follows:

1) A novel Depth Foundation Model-based method, termed DFM-MVS, is proposed to exploit the effective depth prior for MVS in the absence of real-world labels.

2) A depth prior-based pseudo-supervised training mechanism is developed to simulate realistic stereo correspondences using the depth prior, thereby providing effective supervision signals for the MVS network.

3) To mitigate the error propagation problem, a depth prior-guided error correction strategy is designed to refine the mispredicted regions at the coarse scale of the MVS network under the guidance of the depth prior.

4) Experimental results on DTU and Tanks \& Temples datasets show that the proposed DFM-MVS significantly surpasses existing methods without using real-world labels.

\section{Related Work}
\label{sec2}

%-------------------------------------------------------------------------
\subsection{MVS Methods Using Real-World Labels}
\label{sec2.1}

The advent of large-scale 3D datasets has substantially promoted the development of learning-based MVS methods. Most of these methods leverage real-world depth labels derived from datasets to train a well-designed MVS network, thus enabling precise depth estimation and 3D reconstruction. Initially, MVSNet~\cite{mvsnet} established a classical network pipeline, which encodes camera parameters and 2D features into a cost volume through differentiable homography warping, and regularizes the cost volume using 3D CNNs for depth prediction. Despite its progress, MVSNet faces challenges in performance, memory consumption, and time efficiency. To address these limitations, subsequent research has focused on optimizing the pipeline via various approaches. For instance, recurrent methods, like R-MVSNet~\cite{rmvsnet} and AA-RMVSNet~\cite{aarmvsnet}, replace 3D CNNs with RNNs to regularize the cost volume in a sequential manner, significantly reducing memory consumption. To improve performance and time efficiency, coarse-to-fine methods, such as CasMVSNet~\cite{casmvsnet}, CVP-MVSNet~\cite{cvpmvsnet}, and GBi-Net~\cite{gbinet}, are developed to predict a coarse depth map with low resolution and iteratively refine it, finally obtaining a high-quality depth map with high resolution. Built upon the coarse-to-fine network structure, subsequent works have made further enhancements by introducing Transformer modules~\cite{mvstr,transmvsnet,mvster}, mining geometric cues~\cite{geomvsnet,gomvs}, exploiting other representations~\cite{unimvsnet,mvsformer}, and so on. Despite these advancements, the reliance on real-world depth labels remains a significant limitation for the above MVS methods.

%-------------------------------------------------------------------------
\subsection{MVS Methods without Real-World Labels}
\label{sec2.2}

\begin{figure*}
	\centering
	\includegraphics[width=1\linewidth]{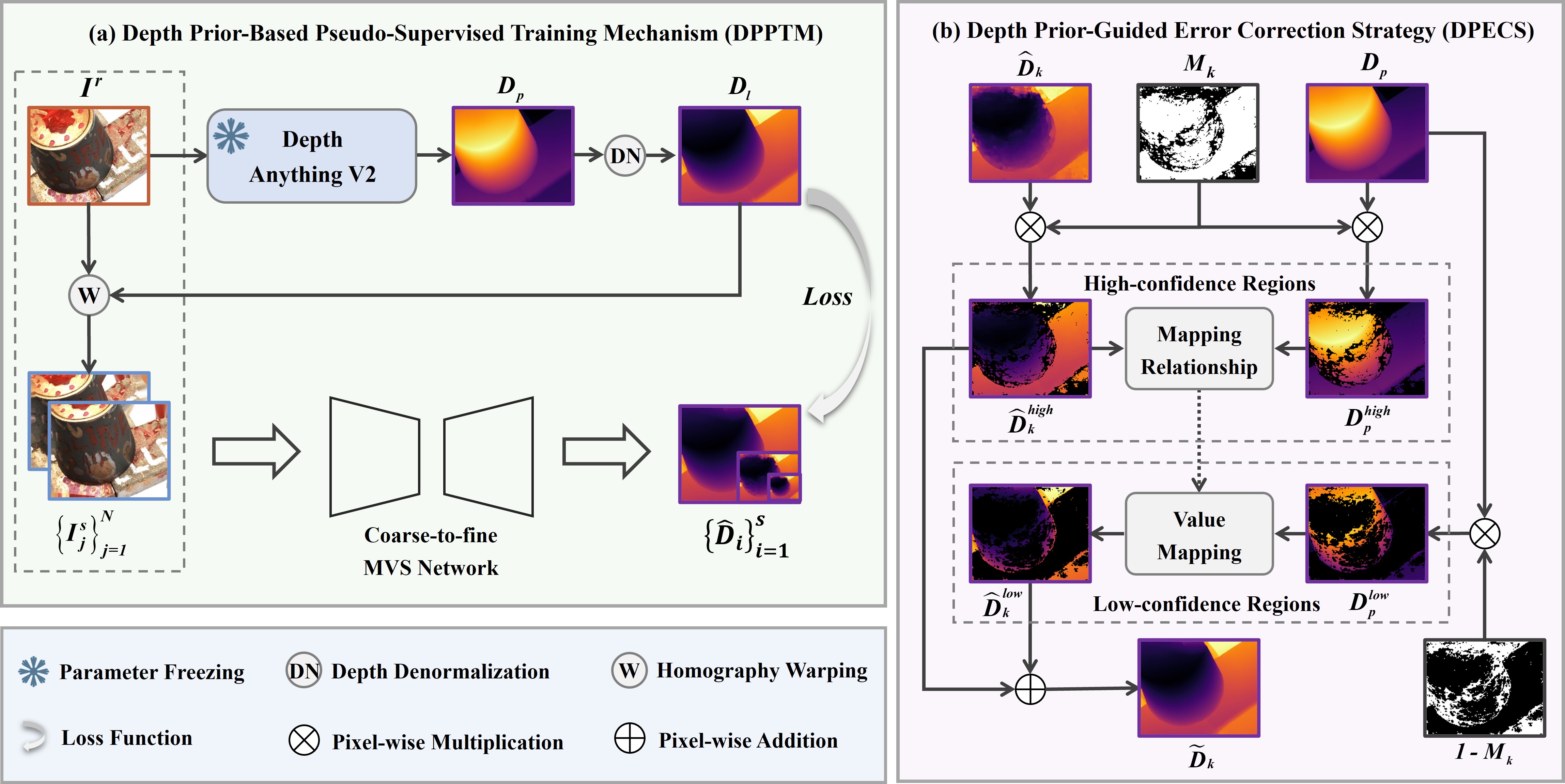}
	\caption{The overall architecture of the proposed DFM-MVS, which includes (a) depth prior-based pseduo-supervised training mechanism (DPPTM), and (b) depth prior-guided error correction strategy (DPECS).}
	\label{overview}
	\vspace{-0.3cm}
\end{figure*}

To eliminate the reliance on real-world depth labels, recent MVS methods have begun to explore alternative training strategies. As a pioneer, Khot \etal~\cite{unsupmvs} transformed MVS into an image reconstruction problem, and proposed a robust multi-view photometric consistency constraint for training. In this method, the source images are first warped to the reference view based on outputs of networks, and then the photometric consistency constraint is enforced by minimizing the difference between the original and warped images. However, the photometric consistency constraint has been proved ineffective in challenging environments, such as occlusions, reflections, and repeated textures~\cite{jcsda, kdmvs, rcmvsnet}. 

To further improve the reliability of the above method, recent researches have explored more effective strategies, which can be mainly divided into two categories. One of the categories introduces some additional constraints based on the photometric consistency constraint. For example, Xu \etal~\cite{jcsda} proposed the data augmentation and cross-view semantics consistency constraints to provide extra priors. Ding \etal~\cite{kdmvs} proposed to use internal features of the MVS network to construct the featuremetric consistency constraint as an additional supervisory signal. Moreover, the integration of other constraints including normal-depth consistency~\cite{m3vsnet}, rendering consistency~\cite{rcmvsnet}, reference-source depth consistency~\cite{ds-mvsnet}, and contrastive consistency~\cite{clmvs} have further enhanced the reliability of training. Instead, another category completely discards the photometric consistency constraint. Concretely, Zhu \etal~\cite{ssc-mvs} proposed SSC-MVS, which first employs a superpixel segmentation algorithm to partition the reference image into distinct regions and assigns random depth values to each region, thus generating a pseudo depth. The pseudo depth is then used to warp the reference image to multiple source views and subsequently constrain the output of the MVS network when inputting the reference image and the warped source images. In this way, the MVS network is trained without real-world depth labels while avoiding the problems of the photometric consistency constraint. Nevertheless, the manually-designed pseudo depth suffers from unsatisfactory quality, which restricts the accuracy and applicability of this method in practical scenarios.

%-------------------------------------------------------------------------
\subsection{Depth Foundation Models}
\label{sec2.3}

Nowadays, the field of computer vision is undergoing a revolution with the rise of foundation models~\cite{vfmsurvey}. These foundation models, pretrained on large-scale datasets capable of encompassing a wide range of data distributions, demonstrate strong zero-shot performance in various real-world scenarios. Notably, the well-known CLIP~\cite{clip}, Segment Anything (SAM)~\cite{sam}, and Stable Diffusion~\cite{stablediff} have already become powerful tools for zero-shot classification, segmentation, and image generation, respectively. Beyond these models, depth foundation models have also emerged due to the fundamental role of monocular depth estimation in modern applications, including robotics, virtual reality, and AI-generated content. For example, MiDaS~\cite{midas} leveraged a combination of labeled datasets to drive early advancements. However, its limited data coverage led to poor performance in some certain scenarios. Depth Anything~\cite{dav1} proposed to tackle this by scaling-up data with massive, cheap, and diverse unlabeled images, exhibiting stronger zero-shot capability. Built upon Depth Anything, Depth Anything V2~\cite{dav2} replaced all labeled real images with synthetic images to seek for better annotation, yielding much powerful zero-shot ability. Inspired by this, this paper leverages the depth prediction from Depth Anything V2 as the effective depth prior, so as to boost MVS in the absence of real-world labels.

\section{Method}
\label{sec3}

%-------------------------------------------------------------------------
\subsection{Overview of DFM-MVS}
\label{sec3.1}

The overall architecture of the proposed depth foundation model-based MVS method (DFM-MVS) is illustrated in \cref{overview}. Firstly, a depth prior-based pseudo-supervised training mechanism (DPPTM) is  proposed to train MVS network in the absence of real-world labels. As shown in \cref{overview} (a), given an RGB image $I^r$, the depth foundation model (\ie, Depth Anything V2) is leveraged to generate the depth prior $D_p$, which demonstrates excellent properties characterizing the real-world depth. After conducting denormalization from $D_p$ to $D_l$, the RGB image $I^r$ is warped to $N$ views using $D_l$, yielding $N$ warped images $\{I_{j}\}{_{j=1}^{N}}$. By setting $I^r$, $\{I_{j}\}{_{j=1}^{N}}$, and $D_l$ as the reference image, source images, and the corresponding supervision signal respectively, a set of MVS training sample is constructed to train a coarse-to-fine MVS network without relying on any real-world label. Besides, to mitigate the error propagation problem inherent in the widely-used coarse-to-fine network structure, a depth prior-guided error correction strategy (DPECS) is further proposed. As shown in \cref{overview} (b), given a coarse depth prediction $\widehat{D}_k$ at the $k^{th}$ scale together with the corresponding confidence mask $M_k$, the depth prior $D_p$ is served as the guidance to refine mispredicted regions in $\widehat{D}_k$, resulting in a refined depth $\widetilde{D}_k$ for subsequent processing at the finer scale. With the proposed DPPTM and DPECS, the prior knowledge embedded in the depth foundation model is effectively exploited, thus boosting MVS without real-world labels. 

%-------------------------------------------------------------------------
\subsection{Depth \hspace{-0.05em}Prior-Based \hspace{-0.05em}Pseudo-Supervised \hspace{-0.05em}Training}
\label{sec3.2}

Motivated by the powerful relative depth estimation ability of Depth Anything V2, which is trained without real-world labels, a depth prior-based pseudo-supervised training mechanism is proposed to leverage Depth Anything V2 to simulate realistic stereo correspondences, thereby facilitating effective supervision for the MVS network.

\cref{overview} (a) shows the proposed depth prior-based pseudo-supervised training mechanism. Given an RGB image $I^r$ sampled from an unlabeled MVS dataset, Depth Anything V2 is first employed to generate the depth prior $D_p$, which is in the inverse depth space with values normalized between 0 and 1. Then, to achieve real view transformations during the subsequent homography warping, $D_p$ is denormalized to the regular depth space to obtain $D_l$, which is defined as: 
\vspace{-0.2cm}
\begin{equation}
\vspace{-0.1cm}
D_l=\frac{1}{D_p\left(\frac{1}{d_{\min}+\eta_1}-\frac{1}{d_{\max}+\eta_2}\right)+\frac{1}{d_{\max}+\eta_2}}+\eta_3
\label{eq1}
\end{equation}
where $d_{\min}$ and $d_{\max}$ denote the minimum and maximum depth respectively. It is worth noting that since MVS is the subsequent process of Structure from Motion (SfM), the camera intrinsic and extrinsic parameters, sparse point clouds, as well as the maximum and minimum depths $d_{\min}$ and $d_{\max}$ derived from SfM are provided in the unlabeled MVS dataset. Besides, $\eta_1$, $\eta_2$, and $\eta_3$ denote random perturbations, in which $\eta_1$ and $\eta_2$ are uniformly sampled from $[0, 0.5(d_{\max}-d_{\min})]$ and $[-0.5(d_{\max}-d_{\min}), 0]$, respectively. Further, $\eta_3$ is uniformly sampled from $[-\eta_1, -\eta_2]$ to ensure $D_l$ within the range $[d_{\min}$, $d_{\max}]$.

Afterwards, with the generated $D_l$, the RGB image $I^r$ is warped to $N$ views, resulting in $N$ warped images $\{I_{j}\}{_{j=1}^{N}}$. In practice, camera poses of warped images are sampled from the adjacent views around $I^r$ according to camera extrinsic parameters provided by the unlabeled MVS dataset. By setting $I^r$, $\{I_{j}\}{_{j=1}^{N}}$, and $D_l$ as the reference image, source images, and the corresponding supervision signal respectively, a set of MVS training sample is constructed, where the pixel-wise multi-view correspondences among $I^r$ and $\{I_{j}\}{_{j=1}^{N}}$ are accurately characterized through $D_l$. 

Subsequently, $I^r$ and $\{I_{j}\}{_{j=1}^{N}}$ are fed into the coarse-to-fine MVS network to generate multi-scale depth predictions $\{\widehat{D}_k\}{_{k=1}^{S}}$, where $S$ denotes the number of scales. With $D_l$ serving as the optimization target, the network is optimized by minimizing a consistency loss between $\{\widehat{D}_k\}{_{k=1}^{S}}$ and $D_l$. The loss function is defined as:
\vspace{-0.3cm}
\begin{equation}
\vspace{-0.3cm}
\boldsymbol{Loss}=\sum_{k=1}^S \lambda_k L_k
\label{eq2}
\end{equation}
where $L_k$ denotes the loss at the $k^{th}$ scale and $\lambda_k$ denotes the corresponding loss weight. 

Since the depth prior exhibits properties characterizing the real-world depth, the constructed training samples closely resemble ones collected with real RGB cameras and 3D scanners. Meanwhile, random perturbations introduced during the denormalization process improve diversity of these samples. Consequently, the MVS network is enabled to be effectively trained with supervision signals derived from these constructed training samples.

%-------------------------------------------------------------------------
\subsection{Depth Prior-Guided Error Correction}
\label{sec3.3}

Existing state-of-the-art MVS methods typically adopt the coarse-to-fine network structure for enhanced performance. However, this structure inherently suffers from the error propagation problem, that prediction errors at the coarse scale may propagate to subsequent scales, impacting final results. Considering this, a depth prior-guided error correction strategy is proposed to leverage the generated depth prior to guide the refinement of the mispredicted regions at the coarse scale, so as to alleviate the error propagation problem.

As shown in \cref{overview} (b), the inputs for the depth prior-guided error correction strategy include a coarse-scale depth prediction $\widehat{D}_k$, its corresponding confidence mask $M_k$, and the depth prior $D_p$. In particular, $M_k$ is generated from the photometric confidence map $\mathcal{P}_k$ at the $k^{th}$ scale of the MVS network, which is formulated as:
\begin{equation}
\vspace{-0.1cm}
M_k(x)=\left\{\begin{array}{l}
1, \text { if } \mathcal{P}_k(x)>\tau \\
0, \text { otherwise }
\end{array}\right\}
\label{eq3}
\end{equation}
where $\mathcal{P}_k$ is a general output of the MVS network, which is often used for depth filtering during merging depths into the point cloud~\cite{mvsnet, casmvsnet, rcmvsnet}. $\mathcal{P}_k(x)$ denotes the photometric confidence value of the corresponding pixel $x$ in $\widehat{D}_k$, and $\tau$ denotes a threshold used to convert $\mathcal{P}_k$ to a binary confidence mask $M_k$. In $M_k$, pixels marked as 0 indicate potential errors in their corresponding depth predictions and need to be corrected. 
To correct these potential errors, $M_k$ is first multiplied by $\widehat{D}_k$  and $D_p$ respectively, resulting in the high-confidence masked depth prediction $\widehat{D}_k^{high}$ and the corresponding high-confidence masked depth prior $D_p^{high}$. Subsequently, a mapping relationship between the depth prior and predicted depth is established by solving the linear function between $D_p^{high}$ and $1/\widehat{D}_k^{high}$. The established mapping relationship is defined as:
\vspace{-0.1cm}
\begin{equation}
\vspace{-0.1cm}
\frac{1}{\widehat{D}_k^{high}(x)}=a_w * D_p^{high}(x)+b_w
\label{eq4}
\end{equation}
where $a_w$ and $b_w$ are the solved coefficients. Based on this mapping relationship, the refinement of depth predictions in low-confidence regions is enabled with the guidance of the depth prior $D_p$. Specifically, the inverse of $M_k$ is taken and multiplied by $D_p$, yielding the low-confidence masked depth prior $D_p^{low}$. By substituting $D_p^{high}$ with $D_p^{low}$ in \cref{eq4}, the refined low-confidence depth prediction $\widetilde{D}_k^{low}$ is obtained, which is formulated as:
\begin{equation}
\widetilde{D}_k^{low}(x)=\frac{1}{a_w * D_p^{low}(x)+b_w}
\label{eq5}
\end{equation}
Finally, by adding the high-confidence masked depth prediction $\widehat{D}_k^{high}$ and refined low-confidence depth prediction $\widetilde{D}_k^{low}$, a complete refined depth prediction $\widetilde{D}_k$ of the $k^{th}$ scale is obtained, which is then served for subsequent processing at the finer ${(k+1)}^{th}$ scale.

With the proposed depth prior-guided error correction strategy, the low-confidence regions of the coarse-scale depth predictions are refined under the guidance of the depth prior generated from the depth foundation model. In this way, errors that may occur in the early scale are corrected immediately, alleviating the error propagation problem and thus helping to improve final performance.

%-------------------------------------------------------------------------
\subsection{Implementation Details}
\label{sec3.4}

Following the common and effective practice in ~\cite{kdmvs, clmvs, ssc-mvs}, the three-scale CasMVSNet~\cite{casmvsnet} is adopted as the coarse-to-fine MVS network. During the training process, the randomly initialized CasMVSNet is first optimized using the proposed depth prior-based pseudo-supervised training mechanism for 8 epochs. In each scale, the loss weight $\lambda_k$ is set as 1. The optimizer is set to Adam with $\beta_{1}=0.9$ and $\beta_{2}=0.999$. The learning rate is initially set as 0.001 and reduced by half at the $4^{th}$, $6^{th}$, and $7^{th}$ epochs. Based on the trained network, the consistency-based training mechanism in~\cite{ssc-mvs} is introduced for further finetuning. Afterwards, the proposed depth prior-guided error correction strategy is integrated into the fintuned network during the testing process to improve final performance. All the above process is implemented using PyTorch on a machine with two NVIDIA GeForce RTX 3090 GPUs. 
\begin{table}[htbp]
	\centering
	\resizebox{1\linewidth}{!}{
		\renewcommand{\arraystretch}{1.25} 
		\normalsize 
		\begin{tabular}{cccccc}
			\Xhline{1px}
			\multicolumn{2}{c}{Methods} & L. & Acc.(mm) & Comp.(mm) & \textbf{Overall(mm)} \\
			\midrule
			\multicolumn{1}{c}{\multirow{3}[1]{*}{\begin{sideways}Traditional\end{sideways}}} 
			& Camp~\cite{camp} & - & 0.835  & 0.554  & 0.695  \\
			& Gipuma~\cite{gipuma} & - & 0.283  & 0.873  & 0.578  \\
			& COLMAP~\cite{colmap} & - & 0.400  & 0.644  &\textbf{0.532} \\
			\hline
			\multirow{23}[-10]{*}{\begin{sideways}Learning-based\end{sideways}} 
			& MVSNet~\cite{mvsnet} & \cmark & 0.396  & 0.527  & 0.462  \\
			& R-MVSNet~\cite{rmvsnet} & \cmark & 0.383  & 0.452  & 0.417  \\
			& CasMVSNet~\cite{casmvsnet} & \cmark & 0.325  & 0.385  & 0.355  \\
			& CVP-MVSNet~\cite{cvpmvsnet} & \cmark & 0.296  & 0.406  & 0.351  \\
			& UniMVSNet~\cite{unimvsnet} & \cmark & 0.352  & 0.278  & 0.315  \\
			& TransMVSNet~\cite{transmvsnet} & \cmark & 0.321  & 0.289  & 0.305  \\
			& GeoMVSNet~\cite{geomvsnet} & \cmark & 0.331  & 0.259  & 0.295  \\
			& MVSFormer++~\cite{mvsformer++} & \cmark & 0.309  & 0.252  & \textbf{0.281} \\
			& GoMVS~\cite{gomvs} & \cmark & 0.347  & 0.227  & 0.287  \\
			\cmidrule{2-6}
			& Unsup\_MVS~\cite{unsupmvs} & \xmark & 0.881  & 1.073  & 0.977  \\
			& $\text{MVS}^{2}$~\cite{mvs2}  & \xmark & 0.760  & 0.515  & 0.637  \\
			& $\text{M}^3\text{VSNet}$~\cite{m3vsnet} & \xmark & 0.636  & 0.531  & 0.583  \\
			& JDACS-MS~\cite{jcsda} & \xmark & 0.398  & 0.318  & 0.358  \\
			& Self-sup CVP~\cite{selfsup-cvp} & \xmark & 0.308  & 0.418  & 0.363  \\
			& AGG-CVCNet~\cite{aggcvcnet} & \xmark & 0.343  & 0.348  & 0.345  \\
			& U-MVS~\cite{umvs} & \xmark & 0.354  & 0.354  & 0.354  \\
			& RC-MVSNet~\cite{rcmvsnet} & \xmark & 0.396  & 0.295  & 0.345  \\
			& KD-MVS~\cite{kdmvs} & \xmark & 0.359  & 0.295  & 0.327  \\
			& DS-MVSNet~\cite{ds-mvsnet} & \xmark & 0.374  & 0.347  & 0.361  \\
			& CL-MVSNet~\cite{clmvs} & \xmark & 0.375  & 0.283  & 0.329  \\
			& SSC-MVS~\cite{ssc-mvs} & \xmark & 0.417  & 0.267  & 0.342  \\
			\cmidrule{2-6}          
			& \textbf{DFM-MVS (Ours)} & \xmark & 0.362  & 0.272  & \textbf{0.317} \\
			\Xhline{1px}
		\end{tabular}
	}
	\vspace{-0.2cm}
	\caption{Quantitative comparison of different methods on the DTU evaluation set (lower is better). "L." represents whether the corresponding method uses real-world labels for training.}
	\label{dtu-tab}%
	\vspace{-0.2cm}
\end{table}%\\

\section{Experimental Results}
\label{sec4}

%-------------------------------------------------------------------------
\subsection{Experimental Settings}
\label{sec4.1}

In this paper, experiments are conducted on commonly-used MVS datasets, including DTU~\cite{dtu} and Tanks \& Temples~\cite{t&t}. DTU is an indoor dataset consisting of 128 scenes, each captured from 49 or 64 camera views under 7 distinct lighting conditions. Tanks \& Temples is a more challenging benchmark dataset with varied lighting and scene complexity across both indoor and outdoor environments. This dataset is further divided into an intermediate subset containing 8 scenes and an advanced subset containing 6 scenes. For both DTU and Tanks \& Temples, camera parameters including intrinsic parameters, extrinsic parameters, minimum and maximum depths are provided through the pre-conducted calibration process~\cite{mvsnet}.

\begin{figure*}
	\centering
	\includegraphics[width=1\linewidth]{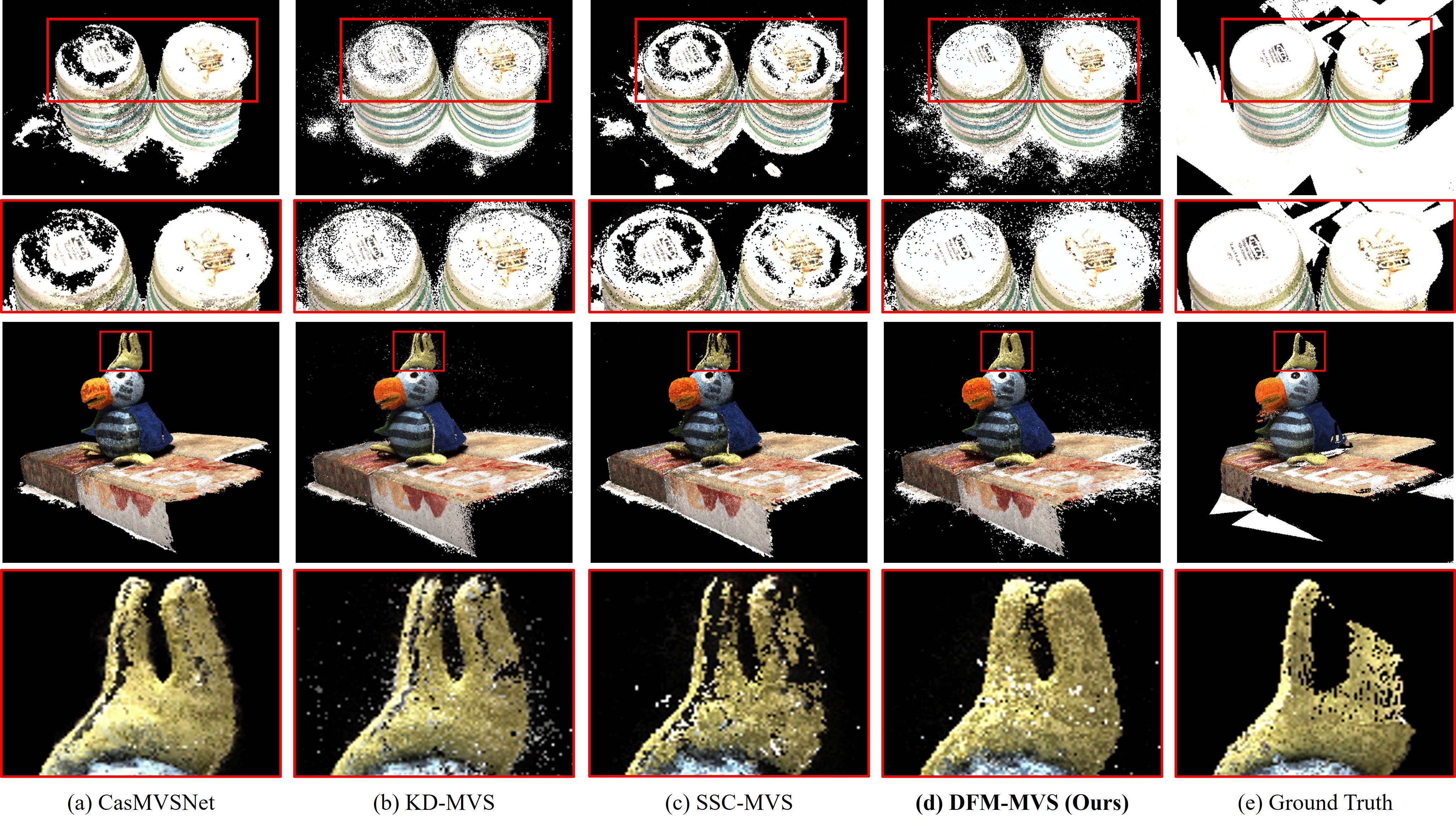}
	\vspace{-0.8cm}
	\caption{Visualization comparison of point clouds reconstructed by different methods on the DTU evaluation set. Rows one and three show the full reconstructed point clouds, while rows two and four show the zoomed-in views of those red-outlined regions.}
	\label{dtu-fig}
	\vspace{-0.6cm}
\end{figure*}

Following the common practice~\cite{rcmvsnet, clmvs, ssc-mvs}, the proposed method is trained on the DTU training set and evaluated on the DTU evaluation set. Specifically, during training, two images are warped from the reference image with a resolution of $640\times512$. During evaluation, one reference image and four source images with a resolution of $1,600\times1,184$ are severed as inputs of the network, where $\tau$ is set as 0.5 for the depth prior-guided error correction strategy. With depths predicted from the network, the depth filtering and fusion method following~\cite{rcmvsnet, clmvs, ssc-mvs} is used to generate the final point cloud. To evaluate the performance of point cloud reconstruction, the distance metrics including accuracy (acc.), completeness (comp.) and overall are reported, where accuracy measures the mean absolute distance from the reconstructed point cloud to ground truth, completeness measures the reverse distance, and overall is the average of accuracy and completeness.

For evaluation on the Tanks \& Temples benchmark dataset, the model trained on the DTU training set is further finetuned on the BlendedMVS training set~\cite{blendedmvs}, which is similar to~\cite{kdmvs}. Specifically, during finetuning, four images are warped from the reference image to combine a 5-view input with the original resolution of $768\times576$. During evaluation, 11 images at the 2K resolution are used to infer depth maps from the network, where $\tau$ is set as 0.7 for the depth prior-guided error correction strategy. The inferred depths are filtered and fused with the dynamic fusion strategy~\cite{d2hcrmvsnet} to reconstruct point clouds. The reconstructed results are uploaded to the official evaluation website to report the F-Score metric, which combines precision and recall. 

%-------------------------------------------------------------------------
\subsection{Experimental Results on DTU}
\label{sec4.2}

Comparison experiments are conducted on the DTU evaluation set to show the performance of different methods. \cref{dtu-tab} reports the quantitative results, where previous traditional, learning-based methods including those using and not using real-world labels are compared with the proposed DFM-MVS. As observed, among learning-based methods without using real-world labels, the proposed DFM-MVS achieves the best overall performance. Besides, DFM-MVS also outperforms its counterpart CasMVSNet~\cite{casmvsnet} that rely on real-world labels, further verifying its superiority.

\begin{table*}[htbp]
	\centering
	\resizebox{1\linewidth}{!}{
		\renewcommand{\arraystretch}{1.25} 
		\normalsize 
		\begin{tabular}{cc|c|ccccccccc|ccccccc}
			\Xhline{1px}
			\multicolumn{2}{c|}{\multirow{2}[4]{*}{Method}} & \multirow{2}[0]{*}{L.} & \multicolumn{9}{c|}{ Intermediate }                                   & \multicolumn{7}{c}{Advanced} \\
			\cline{4-19}    \multicolumn{2}{c|}{} &  & \textbf{Mean} & Fam.  & Fra.  & Hor.  & Lig.  & M60   & Pan.  & Pla.  & Tra.  & \textbf{Mean} & Aud.  & Bal.  & Cou.  & Mus.  & Pal.  & Tem. \\
			\hline
			\multirow{3}[1]{*}{\begin{sideways}Traditional\end{sideways}} 
			& COLMAP~\cite{colmap} & - & 42.14  & 50.41  & 22.25  & 25.63  & 56.43  & 44.83  & 46.97  & 48.53  & 42.04  & 27.24  & 16.02  & 25.23  & 34.70  & 41.51  & 18.05  & 27.94  \\
			& ACMH~\cite{acmh}  & - & 54.82  & 69.99  & 49.45  & 45.12  & 59.04  & 52.64  & 52.37  & 58.34  & 51.61  & 33.73  & 21.69  & 32.56  & 40.62  & 47.27  & 24.04  & 36.17  \\
			& ACMMP~\cite{acmmp} & - & \textbf{59.38} & 70.93  & 55.39  & 51.80  & 63.83  & 55.94  & 59.47  & 59.51  & 58.20  & \textbf{37.84} & 30.05  & 35.36  & 44.51  & 50.95  & 27.43  & 38.73  \\
			\hline 
			\multirow{22}[-5]{*}{\begin{sideways}Learning-based\end{sideways}} 
			& MVSNet~\cite{mvsnet} & \cmark & 43.48  & 55.99  & 28.55  & 25.07  & 50.79  & 53.96  & 50.86  & 47.90  & 34.69  & -     & -     & -     & -     & -     & -     & - \\
			& R-MVSNet~\cite{rcmvsnet} & \cmark & 48.40  & 69.96  & 46.65  & 32.59  & 42.95  & 51.88  & 48.80  & 52.00  & 42.38  & 24.91  & 12.55  & 29.09  & 25.06  & 38.68  & 19.14  & 24.96  \\
			& CasMVSNet~\cite{casmvsnet} & \cmark & 56.84  & 76.37  & 58.45  & 46.26  & 55.81  & 56.11  & 54.06  & 58.18  & 49.51  & 31.12  & 19.81  & 38.46  & 29.10  & 43.87  & 27.36  & 28.11  \\
			& CVP-MVSNet~\cite{cvpmvsnet} & \cmark & 54.03  & 76.50  & 47.74  & 36.34  & 55.12  & 57.28  & 54.28  & 57.43  & 47.54  & -     & -     & -     & -     & -     & -     & - \\
			& UniMVSNet~\cite{unimvsnet} & \cmark & 64.36  & 81.20  & 66.43  & 53.11  & 64.36  & 66.09  & 64.84  & 62.23  & 57.53  & 38.96  & 28.33  & 44.36  & 39.74  & 52.89  & 33.80  & 34.63  \\
			& TransMVSNet~\cite{transmvsnet} & \cmark & 63.52  & 80.92  & 65.83  & 56.94  & 62.54  & 63.06  & 60.00  & 60.20  & 58.67  & 37.00  & 24.84  & 44.59  & 34.77  & 46.49  & 34.69  & 36.62  \\
			& GeoMVSNet~\cite{geomvsnet} & \cmark & 65.89  & 81.64  & 67.53  & 55.78  & 68.02  & 65.49  & 67.19  & 63.27  & 58.22  & 41.52  & 30.23  & 46.53  & 39.98  & 53.05  & 35.98  & 43.34  \\
			& MVSFormer++~\cite{mvsformer++} & \cmark & \textbf{67.18} & 82.69  & 69.44  & 64.24  & 69.16  & 64.13  & 66.43  & 61.19  & 60.12  & 41.60  & 29.93  & 45.69  & 39.46  & 53.58  & 35.56  & 45.39  \\
			& GoMVS~\cite{gomvs} & \cmark & 66.44  & 82.68  & 69.23  & 69.19  & 63.56  & 65.13  & 62.10  & 58.81  & 60.80  & \textbf{43.07} & 35.52  & 47.15  & 42.52  & 52.08  & 36.34  & 44.82  \\
			\cline{2-19}
			& $\text{MVS}^{2}$~\cite{mvs2}  & \xmark & 37.21  & 47.74  & 21.55  & 19.50  & 44.54  & 44.86  & 46.32  & 43.38  & 29.72  & -     & -     & -     & -     & -     & -     & - \\
			& $\text{M}^3\text{VSNet}$~\cite{m3vsnet} & \xmark & 37.67  & 47.74  & 24.38  & 18.74  & 44.42  & 43.55  & 44.95  & 47.39  & 30.31  & -     & -     & -     & -     & -     & -     & - \\
			& JDACS-MS~\cite{jcsda} & \xmark & 45.48  & 66.62  & 38.25  & 36.11  & 46.12  & 46.66  & 45.25  & 47.69  & 37.16  & -     & -     & -     & -     & -     & -     & - \\
			& Self-sup CVP~\cite{selfsup-cvp} & \xmark & 46.71  & 64.95  & 38.79  & 24.98  & 49.73  & 52.57  & 51.53  & 50.66  & 40.45  & -     & -     & -     & -     & -     & -     & - \\
			& AGG-CVCNet~\cite{aggcvcnet} & \xmark & 57.81  & 77.39  & 59.74  & 50.68  & 54.02  & 59.59  & 56.04  & 54.42  & 50.57  & 31.41  & 22.11  & 36.50  & 30.05  & 39.88  & 27.94  & 31.95  \\
			& U-MVS~\cite{umvs} & \xmark & 57.15  & 76.49  & 60.04  & 49.20  & 55.52  & 55.33  & 51.22  & 56.77  & 52.63  & 30.97  & 22.79  & 35.39  & 28.90  & 36.70  & 28.77  & 33.25  \\
			& RC-MVSNet~\cite{rcmvsnet} & \xmark & 55.04  & 75.26  & 53.50  & 45.52  & 53.49  & 54.85  & 52.30  & 56.06  & 49.37  & 30.82  & 21.72  & 37.22  & 28.62  & 37.37  & 27.88  & 32.09  \\
			& KD-MVS~\cite{kdmvs} & \xmark & 64.14  & 80.42  & 67.42  & 54.02  & 64.52  & 64.18  & 61.60  & 62.37  & 58.59  & 37.96  & 27.24  & 44.10  & 35.47  & 49.16  & 34.68  & 37.11  \\
			& DS-MVSNet~\cite{ds-mvsnet} & \xmark & 54.76  & 74.99  & 59.78  & 42.15  & 53.66  & 53.52  & 52.57  & 55.38  & 46.03  & -     & -     & -     & -     & -     & -     & - \\
			& CL-MVSNet~\cite{clmvs} & \xmark & 59.39  & 76.35  & 62.37  & 49.93  & 60.02  & 57.44  & 59.97  & 56.74  & 52.28  & 37.03  & 28.07  & 43.55  & 37.47  & 50.86  & 31.45  & 30.78  \\
			& SSC-MVS~\cite{ssc-mvs} & \xmark & 57.26  & 70.70  & 57.98  & 46.22  & 62.30  & 53.36  & 56.06  & 57.55  & 53.94  & 34.89  & 21.43  & 40.74  & 35.15  & 45.17  & 31.22  & 35.58  \\
			\cline{2-19}          & \textbf{DFM-MVS (Ours)} & \xmark & \textbf{64.64} & 80.30  & 68.31  & 58.92  & 66.24  & 64.03  & 62.04  & 61.01  & 56.26  & \textbf{40.35} & 29.32  & 46.28  & 40.66  & 52.52  & 34.49  & 38.83  \\
			\Xhline{1px}
		\end{tabular}%
	}
	\vspace{-0.2cm}
	\caption{Quantitative results of different methods on the Tanks \& Temples benchmark dataset (higher is better). "L." represents whether the corresponding method uses real-world labels for training.}
	\label{tnt-tab}%
	\vspace{-0.3cm}
\end{table*}%

\begin{figure*}
	\centering
	\includegraphics[width=0.99\linewidth]{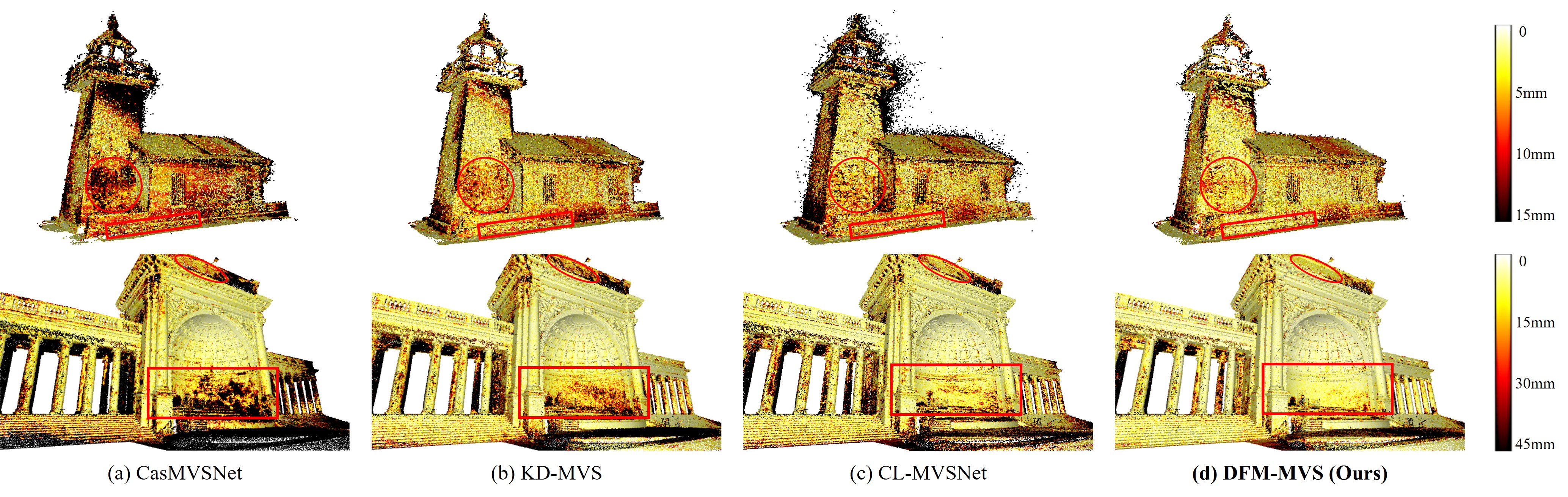}
	\vspace{-0.2cm}
	\caption{Error visualization of point clouds reconstructed by different methods on the Tanks \& Temples benchmark dataset. Darker areas in the map indicate larger errors in the point cloud.}
	\label{tnt-fig}
	\vspace{-0.2cm}
\end{figure*}

The visualization comparison of point clouds reconstructed by different methods is shown in \cref{dtu-fig}. It can be observed that the proposed DFM-MVS significantly improves reconstruction quality over CasMVSNet~\cite{casmvsnet} and other state-of-the-art methods without using real-world labels (\ie, KD-MVS~\cite{kdmvs}, and SSC-MVS~\cite{ssc-mvs}), especially in challenging regions, such as untextured surfaces (\eg, the second row), and thin structures (\eg, the fourth row). This is mainly because the proposed DFM-MVS effectively leverages the depth prior generated from the depth foundation model to construct reliable supervision and correct early-stage errors in the network.

%-------------------------------------------------------------------------
\subsection{Experimental Results on Tanks \& Temples}
\label{sec4.3}

To verify the generalization of the proposed method, comparison experiments are further conducted on the intermediate and advanced subsets of the Tanks \& Temples benchmark dataset. \cref{tnt-tab} reports quantitative results of different methods. As demonstrated, the proposed DFM-MVS achieves the highest mean F-Score among methods not using real-world labels on both the intermediate and advanced subsets. Additionally, the proposed DFM-MVS also exhibits performance on par with state-of-the-art methods that rely on real-world labels across both subsets. 

Moreover, \cref{tnt-fig} shows the error visualization of point clouds reconstructed by different methods on the Tanks \& Temples benchmark dataset. It can be seen that the proposed DFM-MVS achieves lower reconstruction errors, especially in those challenging regions marked by red boxes and circles. These observations further demonstrate the effectiveness of the proposed method.

%-------------------------------------------------------------------------
\subsection{Ablation Study}
\label{sec4.4}
\begin{table}[t]
	\centering
	\resizebox{1\linewidth}{!}{
		\renewcommand{\arraystretch}{1.05} 
		\normalsize 
		\begin{tabular}{cc|ccc}
			\Xhline{1px}
			DPPTM & DPECS & Acc.(mm) & Comp.(mm) & \textbf{Overall(mm)} \\
			\hline
			&            & 0.417  & 0.267  & 0.342  \\
			\checkmark  &            & 0.365  & 0.287  & 0.326  \\
			\checkmark  & \checkmark & 0.362  & 0.272  & \textbf{0.317} \\
			\Xhline{1px}
		\end{tabular}%
	}
	\vspace{-0.1cm}
	\caption{Quantitative results of the ablation study conducted on the DTU evaluation set (lower is better).}
	\label{abl-tab}%
	\vspace{-0.6cm}
\end{table}%

To analyze the effectiveness of the key components in the proposed DFM-MVS, the ablation study is conducted on the DTU evaluation set. The quantitative results are shown in \cref{abl-tab}, where “DPPTM” and “DPECS” denote the proposed depth prior-based pseudo-supervised training mechanism and depth prior-guided error correction strategy, respectively. Note that the method without DPPTM and DPECS means using the manually-designed depth in~\cite{ssc-mvs} instead of the depth prior generated from Depth Anything V2 to construct pseudo-supervision. It can be seen that the method with DPPTM achieves better overall performance compared to the method without both key components. This is mainly because the proposed DPPTM effectively leverages the depth prior generated from Depth Anything V2 to construct more realistic stereo correspondences, thereby improving the effectiveness of pseudo-supervision. Besides, as demonstrated, the overall performance is further enhanced by adding DPECS, which is attributed to the mitigation of error accumulation problem under the guidance of the depth prior.

\begin{figure}
	\centering
	\includegraphics[width=0.88\linewidth]{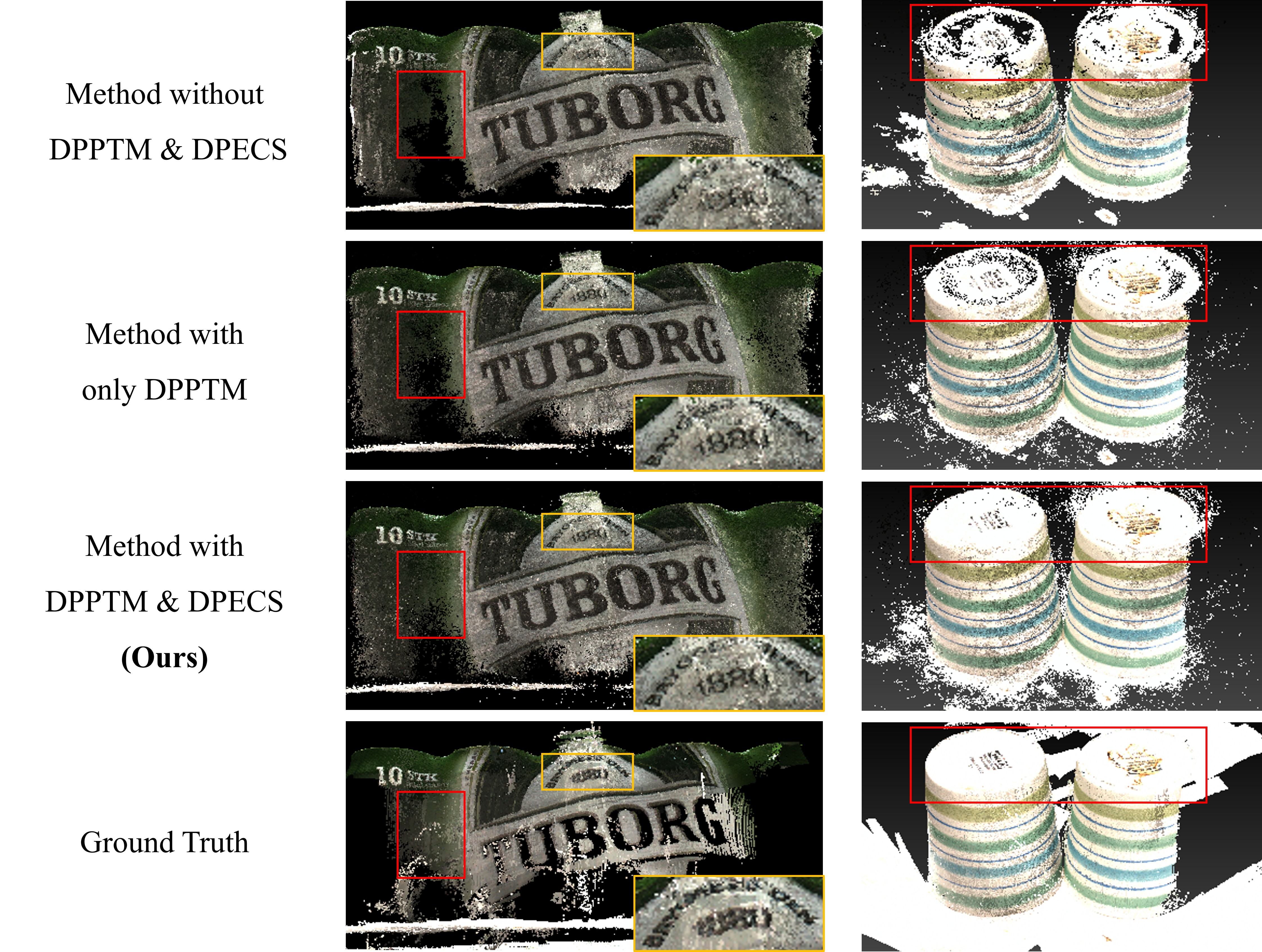}
	\caption{Visualization comparison of point clouds reconstructed by different variant methods on the DTU evaluation set. }
	\label{abl-fig}
	\vspace{-0.6cm}
\end{figure}

\cref{abl-fig} shows the visualization comparison of point clouds reconstructed by different variant methods on the DTU evaluation set. As observed in the second column, compared with the method without DPPTM and DPECS, adding DPPTM significantly improves the reconstruction accuracy (\eg, yellow-outlined regions). Based on this, incorporating DPECS further improves the reconstruction completeness (\eg, red-outlined regions), thus resulting in the superior overall quality. Besides, as shown in the third column, as each component is successively added, the reconstruction result gradually gets closer to the ground-truth point cloud, which further verifies the effectiveness of the proposed depth prior-based pseudo-supervised training mechanism and depth prior-guided error correction strategy.

\subsection{Discussion}
\label{sec4.5}

\paragraph{Evaluation of directly using Depth Anything V2 for reconstrcution.}
Generally, given a depth foundation model, it is an intuitive idea to use the model to estimate depth maps per view and then fuse them to reconstruct point clouds. To evaluate this idea, experiments are conducted on the DTU evaluation set, and results are shown in~\cref{dis1}. Specifically, for the method “D.A.V2”, similar to~\cite{depthreggs}, the outputs from Depth Anything V2 are adjusted to align the scale and shift with SfM sparse points. For the method “D.A.V2-Metric”, the metric depth model fine-tuned on an indoor dataset (\ie, Hypersim~\cite{hypersim}) is used to directly predict real-world absolute depths. Then, both “D.A.V2” and “D.A.V2-Metric” adopt the same cross-view consistency-based filtering and fusion method with the proposed DFM-MVS, to reconstruct point clouds. As observed in~\cref{dis1}, “D.A.V2” suffers from notable depth errors and poor reconstruction performance. Besides, “D.A.V2-Metric” also shows significant depth errors, probably due to distributional differences between Hypersim and DTU. In this case, point clouds fail to be reconstructed for some scans after the cross-view consistency check, making the evaluation metrics invalid. In contrast to the above methods, the proposed DFM-MVS exploits Depth Anything V2 to construct effective pseudo-supervision for MVS, thus yielding state-of-the-art accuracy of multi-view depth estimation and 3D reconstruction performance.
 \begin{table}[t]
 	\centering
 	\resizebox{1\linewidth}{!}{
 		\renewcommand{\arraystretch}{1.01} 
 		\Huge
 		\begin{tabular}{c|cccc}
 			\Xhline{2px}
 			Method & \textless8mm$\uparrow$  & Acc.(mm)$\downarrow$ & Comp.(mm)$\downarrow$ & \textbf{Overall(mm)}$\downarrow$ \\
 			\hline
 			D.A.V2 & 0.516  & 3.717  & 3.389  & 3.351  \\
 			D.A.V2-Metric & 0.015  & /     & /     & / \\
 			\textbf{DFM-MVS (Ours)} & 0.872  & 0.362  & 0.272  & \textbf{0.317} \\
 			\Xhline{2px}
 		\end{tabular}%
 	}
 	\vspace{-0.2cm}
 	\caption{Evaluation of directly using Depth Anything V2 for reconstrcution.\textless8mm means the ratio of depth error within 8mm.}
 	\vspace{-0.2cm}
 	\label{dis1}%
 \end{table}%

 \begin{table}[t]
 	\centering
 	\resizebox{1\linewidth}{!}{
 		\begin{tabular}{c|ccc}
 			\Xhline{1px}
 			Method & Acc.(mm) & Comp.(mm) & \textbf{Overall(mm)} \\
 			\hline
 			DPPTM w/o r.p. & 0.372  & 0.292  & 0.332  \\
 			DPPTM & 0.365  & 0.287  & \textbf{0.326} \\
 			\Xhline{1px}
 		\end{tabular}%
 	}
	\vspace{-0.2cm}
 	\caption{Evaluation of random perturbations introduced in DPPTM. “r.p.” means random perturbations.}
 	\vspace{-0.6cm}
 	\label{dis2}%
 \end{table}%
\vspace{-0.5cm}
\paragraph{Evaluation of random perturbations introduced in DPPTM.} To verify the effectiveness of random perturbations $\eta_1$, $\eta_2$, and $\eta_3$ introduced in DPPTM, experiments are conducted on the DTU evaluation set. ~\cref{dis2} shows the experimental results. As demonstrated, removing random perturbations causes a performance decline in terms of all metrics. This is mainly because the introduced random perturbations  improve diversity of the constructed training samples, thereby enhancing the robustness of the network.
\vspace{-0.5cm}
\paragraph{Comparisons to SSC-MVS.} To generate the pseudo depth, SSC-MVS~\cite{ssc-mvs} relies on a complex procedure involving carefully selected superpixel segmentation algorithms. Nevertheless, the manually-designed pseudo depth suffers from unsatisfactory quality, limiting the effectiveness of pseudo-supervision. In contrast, following the trend of the development of vision foundation models, our DFM-MVS offers a more convenient way to generate more effective pseudo-supervision, where some new attempts such as random perturbations are also proven to be valid. Besides, the proposed DPECS alleviates the inherent error propagation problem of existing coarse-to-fine networks, further providing new insights for MVS.
 \vspace{-0.2cm}
\section{Conclusion}
\label{sec5}
This paper proposes DFM-MVS, a novel method that exploits the depth foundation model to generate the effective depth prior for boosting MVS in the absence of real-world labels. In the proposed DFM-MVS, a depth prior-based pseudo-supervised training mechanism is designed to leverage the generated depth prior to simulate realistic stereo correspondences, so as to provide effective supervision for the MVS network. Further, to mitigate the error propagation problem inherent in the coarse-to-fine network structure, a depth prior-guided error correction strategy is presented to refine early-stage prediction errors under the guidance of the depth prior. Experimental results on widely-used MVS datasets demonstrate the remarkable performance of the proposed DFM-MVS.
{
    \small
    \bibliographystyle{ieeenat_fullname}
    \bibliography{main}
}

\end{document}